# Inference with Causal Independence in the CPSC Network

Nevin Lianwen Zhang
Dept. of Computer Science, University of Science & Technology, Hong Kong
lzhang@cs.ust.hk

## Abstract

This paper reports experiments with the causal independence inference algorithm proposed by Zhang and Poole (1994b) on the CPSC network created by Pradhan et al (1994). It is found that the algorithm is able to answer 420 of the 422 possible zero-observation queries, 94 of 100 randomly generated five-observation queries, 87 of 100 randomly generated ten-observation queries, and 69 of 100 randomly generated twenty-observation queries.

## 1 Introduction

The CPSC network is a multilevel, multivalued medical Bayesian network (BN). It was created by Pradhan et al (1994) based on the Computer-based Patient Case Simulation system (CPSC-PM) developed by Parker and Miller.

The CPSC network is one of the largest BNs in use at the present time. To the best of my knowledge, none of the existing implementations of BN are able to make inference with the network.

The CPSC network contains abundant causal independencies. This makes it a good test case for the inference algorithm proposed by Zhang and Poole (1994b), the theme of which is to exploit causal independencies for efficiency gains. Experiments have been performed. It is found that the algorithm is able to answer 420 of the 422 possible zero-observation queries, 94 of 100 randomly generated five-observation queries, 87 of 100 randomly generated ten-observation queries, and 69 of 100 randomly generated twenty-observation queries. Here, a five-observation query means a query about the posterior probability of one variable given five observations.

In addition to the reporting of experiment results (Section 5), this paper also gives a somewhat new presentation of the algorithm to help the reader in gaining a good understanding of the key issue and of the essence of the algorithm. Terminological and technical modifications are also introduced.

It is well known that conditional independencies lead to the factorization of a joint probability into the multiplication of a list of conditional probabilities. The concept of causal independence we use (Section 2) allows one to further factorize each of those conditional probabilities into a combination of "even-smaller" factors, resulting in a finer-grain factorization of the joint probability (Section 3). The key issue is that factors in this finer-grain factorization usually cannot be combined in arbitrary order (Section 4.2). This difficulty is overcome through a general combination operator (section 4.3), the concept of deputation (Section 4.4), and a constraint on the elimination ordering (Section 4.5).

## 2 Causal independence

*Causal independence* refers to the situation where several causes (or variables) $c_1$, $c_2$, ..., $c_m$ contribute independently to an effect (or variable) $e$. The contribution $\xi_i$ by $c_i$ probabilistically depends on $c_i$ itself and is independent of all other causes given $c_i$. The total contribution that $e$ receives is an combination $e = \xi_1 * \xi_2 * \ldots * \xi_m$ of the individual contributions, where $*$ is a certain associative and commutative binary operator. When it is the case, we call the variable $e$ a *convergent variable* since it is where independent contributions from different sources are collected and combined. The operator $*$ is called the *base combination operator of $e$*.

The noisy-OR gate (Pearl 1988) is an example of causal independence where all the variables are binary and the logic OR operator "∨" is used to combine individual contributions. Pradhan et al (1994) introduce a generalization of the noisy-OR gate model called the noisy-MAX gate and use it extensively in the CPSC network. The noisy-MAX gate is another example of causal independence where the possible values of $e$ are ordered and the "MAX" operator is used to combine individual contributions. Other examples of causal independence include noisy-AND gates and noisy-adders.

In a causal independence model, the conditional probability $P(e|c_1, c_2, \ldots, c_m)$ can be obtained from the conditional probabilities $P(\xi_i|c_i)$. To see this, let us first define a function $f_i(e, c_i)$ for each $i$ as follows: $f_i(e=\alpha, c_i=\beta) =_{def} P(\xi_i=\alpha|c_i=\beta)$ for any value $\alpha$ of $e$ and any value $\beta$ of $c_i$.

We also need to define an operator to combine the $f_i$'s. Let $f(e, A, B)$ and $g(e, A, C)$ be two functions, where $A$, $B$, and $C$ are three lists of variables and $B$ and $C$ do not intersect. The *combination* $f \otimes_* g$ of $f$ and $g$ is defined as follows: for any particular value $\alpha$ of $e$,

$$f \otimes_* g(e=\alpha, A, B, C)$$
$$=_{def} \sum_{\alpha_1 * \alpha_2 = \alpha} f(e=\alpha_1, A, B) \; g(e=\alpha_2, A, C). \quad (1)$$

We shall refer $\otimes_*$ as the *functional combination operator of $e$*. It is important to notice that $*$ combines values of $e$, while $\otimes_*$ combines functions of $e$. One can easily verify that the functional combination operator $\otimes_*$ is also commutative and associative.

It can be shown that the conditional probability $P(e|c_1, c_2, \ldots, c_m)$ of the convergent variable $e$ can be expressed as the combination of the $f_i$'s, i.e.

$$P(e|c_1, \ldots, c_m) = f_1(e, c_1) \otimes_* \ldots \otimes_* f_m(e, c_m). \quad (2)$$

The right hand of the equation makes sense because $\otimes_*$ is commutative and associative. Again, the base combination operator determines how contributions from indifferent sources are combined, while the functional combination operator is the reflection of the base operator in terms of conditional probability.

For example, the conditional probability $P(e|c_1, c_2)$ of the convergent variable $e$ in a noisy-OR gate with causes $c_1$ and $c_2$ is given by

$$P(e=\alpha|c_1, c_2) = \sum_{\alpha_1 \vee \alpha_2 = \alpha} P(\xi_1=\alpha_1|c_1)P(\xi_2=\alpha_2|c_2),$$

where $\xi_i$ is the contribution by $c_i$ and $\alpha$ can be either 0 or 1. Hence,

$$P(e|c_1, c_2) = f_1(e, c_1) \otimes_\vee f_2(e, c_2).$$

It is interesting to notice the similarity between equation (2) and the following property of conditional independence: if a variable $x$ is independent of another variable $z$ given a third variable $y$, then there exist non-negative functions $f(x, y)$ and $g(y, z)$ such that

$$P(x, y, z) = f(x, y)g(y, z). \quad (3)$$

In equation (3) conditional independence allows us to factorize a joint probability into factors that involve less variables, while in equation (2) causal independence allows us to factorize a conditional probability into factors that involve less variables. The only difference lies in the way the factors are combined.

Conditional independence has been used to reduce inference complexity in Bayesian networks (Pearl 1988, Lauritzen and Spiegelhalter 1988 and Jensen *et al* 1990). The rest of this paper investigates how to use causal independence for the same purpose.

## 3 Heterogeneous factorization of joint probabilities

This section discusses factorization of joint probabilities and introduces the concept of heterogeneous factorization (HF).

A fundamental assumption under the theory of probabilistic reasoning is that a joint probability is adequate for capturing experts' knowledge and beliefs relevant to a reasoning task. Factorization and Bayesian networks come into play because joint probability is difficult, if not impossible, to directly assess, store, and reason with.

Let $P(x_1, x_2, \ldots, x_n)$ be a joint probability over variables $x_1, x_2, \ldots, x_n$. By the chain rule of probabilities, we have

$$P(x_1, x_2, \ldots, x_n)$$
$$= P(x_1)P(x_2|x_1) \ldots P(x_n|x_1, \ldots, x_{n-1}). \quad (4)$$

For any $i$, there might be a subset $\pi_i \subseteq \{x_1, \ldots, x_{i-1}\}$ such that $x_i$ is conditionally independent of all the other variables in $\{x_1, \ldots, x_{i-1}\}$ given variables in $\pi_i$, i.e. $P(x_i|x_1, \ldots, x_{i-1}) = P(x_i|\pi_i)$. Equation (4) can hence be rewritten as

$$P(x_1, x_2, \ldots, x_n) = \prod_{i=1}^{n} P(x_i|\pi_i). \quad (5)$$

Equation (5) factorizes the joint probability $P(x_1, x_2, \ldots, x_n)$ into a multiplication of factors $P(x_i|\pi_i)$. While the joint probability involves all the $n$ variables, each of the factors might involve only a small number of variables. This fact implies savings in assessing, storing, and reasoning with probabilities.

A *Bayesian network* is constructed from the factorization as follows: build a directed graph with nodes $x_1$, $x_2, \ldots, x_n$ such that there is an arc from $x_j$ to $x_i$ if and only if $x_j \in \pi_i$, and associate the conditional probability $P(x_i|\pi_i)$ with the node $x_i$. $P(x_1, \ldots, x_n)$ is said to be the *joint probability* of the Bayesian network. Also nodes in $\pi_i$ are called *parents* of $x_i$. The term node will be use interchangeably with the term variable in the rest of the paper.

The factorization given by equation (5) is *homogeneous* in the sense that all the factors are combined in the same way, i.e. by multiplication.

Let $x_{i1}, \ldots, x_{im_i}$ be the parents of $x_i$. If $x_i$ is a convergent variable, then the conditional probability

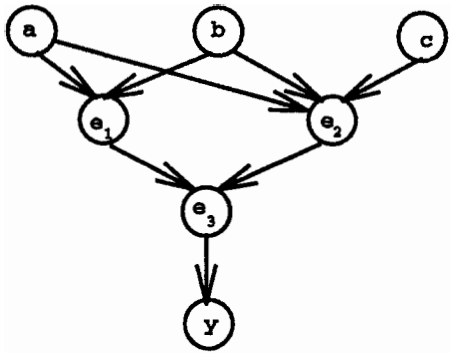

Figure 1: A Bayesian network, where $e_1$, $e_2$, and $e_3$ are convergent variables.

$P(x_i|\pi_i)$ can be further factorized into

$$P(x_i|\pi_i) = f_{i1}(x_i, x_{i1}) \otimes_i \ldots \otimes_i f_{im_i}(x_i, x_{im_i}), \qquad (6)$$

where $\otimes_i$ is the functional combination operator of $x_i$. The fact that $\otimes_i$ might be other than multiplication leads to the concept of heterogeneous factorization. The word heterogeneous reflects the fact that different factors might be combined in different manners.

As an example, consider the Bayesian network in Figure 1. The network states that $P(a,b,c,e_1,e_2,e_3,y)$ can be factorized into a multiplication of $P(a)$, $P(b)$, $P(c)$, $P(e_1|a,b)$, $P(e_2|a,b,c)$, $P(e_3|e_1,e_2)$, and $P(y|e_3)$.

If the $e_i$'s are convergent variables, then the conditional probabilities of the $e_i$'s can be further factorized as follows:

$$\begin{aligned} P(e_1|a,b) &= f_{11}(e_1,a) \otimes_1 f_{12}(e_1,b) \\ P(e_2|a,b,c) &= f_{21}(e_2,a) \otimes_2 f_{22}(e_2,b) \otimes_2 f_{23}(e_2,c) \\ P(e_3|e_1,e_2) &= f_{31}(e_3,e_1) \otimes_3 f_{32}(e_3,e_1) \end{aligned}$$

where the factor $f_{11}(e_1,a)$, for instance, captures the contribution of $a$ to $e_1$, and where the $\otimes_i$ is the functional combination operator of the $e_i$.

The factorization of $P(a,b,c,e_1,e_2,e_3,y)$ into the factors: $P(a)$, $P(b)$, $P(c)$, $P(y|e_3)$, $f_{11}(e_1,a)$, $f_{12}(e_1,b)$, $f_{21}(e_2,a)$, $f_{22}(e_2,b)$, $f_{23}(e_2,c)$, $f_{31}(e_3,e_1)$, and $f_{32}(e_3,e_2)$ is called a *heterogeneous factorization (HF)*. We shall call the $f_{ij}$'s *heterogeneous factors* since they might be combined with other factors by operators other than multiplication. In contrast, we shall say that the factors $P(a)$, $P(b)$, $P(c)$, and $P(y|e_3)$ are *homogeneous*. Since the heterogeneous factorization can be read from the BN in Figure 1, we say that the BN *denotes* the factorization.

## 4 Exploiting causal independencies in inference

The question is how to make use of causal independencies in inference. This section reviews the approach proposed by Zhang and Poole (1994b). Minor technical modifications are introduced.

Let us consider queries of the form $P(X, Y=Y_0)$, where $X$ is a list of interesting variables, $Y$ is a list of observed variables, and $Y_0$ is the corresponding list of observed values. It suffices to only consider such queries because the posterior probability $P(X|Y=Y_0)$ can be readily obtained from $P(X, Y=Y_0)$ and $P(Y=Y_0)$.

### 4.1 Irrelevance

There might be variables in a BN that are irrelevant to a query (Geiger *et al* 1988, Lauritzen *et al* 1990). The paper assumes that all the irrelevant variables have been pruned.

A factor can be represented as a multidimensional array. Portions of the array that represents a factor might also be irrelevant to a query. In a BN, a *regular variable* is one that is not a convergent variable. If a factor $f(y,Z)$ involves regular variable $y$ and $y$ is observed to be $y_0$, then the values in the cells where $y \neq y_0$ are irrelevant.

Note that the same cannot be done if $y$ is a convergent variable. In this case, $f(y,Z)$ is a heterogeneous factor. There might exist other heterogeneous factors that contain $y$. When combining those factors, we might need values of $f(y,Z)$ for cases where $y \neq y_0$ (see equation (1)).

We assume irrelevant portions of all the factor arrays have been pruned and treat the observed regular variables as special variables with only one possible value. Pruning irrelevant portions of factor arrays is especially important when there is a large number of observations.

### 4.2 A difference between homogeneous and heterogeneous factorizations

One way to compute $P(X, Y=Y_0)$ is to sum out the variables outside $X$ one by one. With a homogeneous factorization, summing out one variable $z$ is easy. One can simply remove all the factors that involve $z$ from the list of factors; combine them by multiplication; sum out $z$ from the combination; and put the resulting factor onto the list of factors (Zhang and Poole 1994a). This is essentially what takes place in the well known clique tree propagation algorithm (Lauritzen and Spiegelhalter 1988 and Jensen *et al* 1990). The correctness of doing so is guaranteed by the fact that factors in a homogeneous factorization can be combined in arbitrary order.

Unlike in a homogeneous factorization, factors in a heterogeneous factorization in general cannot be combined in arbitrary order. In our example, summing out variable $a$ requires combining $f_{11}(e_1,a)$ and $f_{21}(e_2,a)$. But by definition $f_{11}$ needs to be combined with $f_{12}$ before being combined with any other factors, including $f_{21}$. Thus, we need more flexibility on the order by which heterogeneous factors can be combined. This is the key issue one needs to address in order to make use of causal independence in inference.

Zhang and Poole (1994b) achieve such flexibility through a general combination operator, the concept of deputation, and a constraint on the order by which variables are summed out.

### 4.3 A general combination operator

Suppose $e_1, \ldots, e_k$ are convergent variables with base combination operator $*_1, \ldots, *_k$. Let $f(e_1, \ldots, e_k, A, B)$ and $g(e_1, \ldots, e_k, A, C)$ be two functions, where the $A$ is a list of regular variables and $B$ and $C$ do not intersect (they can contain convergent as well as regular variables). Then, the *combination* $f \otimes g$ of $f$ and $g$ is defined as follows: for any particular value $\alpha_i$ of $e_i$,

$$f \otimes g(e_1 = \alpha_1, \ldots, e_k = \alpha_k, A, B, C)$$
$$=_{def} \sum_{\alpha_{11} *_1 \alpha_{12} = \alpha_1} \cdots \sum_{\alpha_{k1} *_k \alpha_{k2} = \alpha_2} \quad (7)$$
$$f(e_1 = \alpha_{11}, \ldots, e_k = \alpha_{k1}, A, B)$$
$$g(e_1 = \alpha_{12}, \ldots, e_k = \alpha_{k2}, A, C). \quad (8)$$

A few notes are in order. First, fixing a list of convergent variables and their base combination operators, one can use the operator $\otimes$ to combined two arbitrary functions. Second, since the base combination operators are commutative and associative, the operator $\otimes$ is also commutative and associative.

In the following, we shall work with a given BN, which has a fixed list of convergent variables. Consequently, we can use $\otimes$ to combine any two heterogeneous factors and the heterogeneous factors can be combined in any order. We shall refer to $\otimes$ as the *general combination operator*.

Third, when $k = 1$ equation (7) reduces to equation (1). Finally when $k = 0$, $f \otimes g$ is simply the multiplication of $f$ and $g$.

### 4.4 Deputation

Let $e$ be a convergent node in a BN. To *depute* $e$ is to make a copy $e'$ of $e$, make the children of $e$ to be children of $e'$, make $e'$ a child of $e$, and set the conditional probability $P(e'|e)$ to be as follows:

$$P(e'|e) = \begin{cases} 1 & \text{if } e = e' \\ 0 & \text{otherwise} \end{cases} \quad (9)$$

We shall call $e'$ the *deputy* of $e$. We shall also call $P(e|e')$ the *deputing function* and sometimes write it as $I(e, e')$ since $P(e|e')$ ensures that $e$ and $e'$ be the same.

The BN in Figure 1 becomes the one in Figure 2 after the deputation of all the convergent variables. It is called the *the deputation* of the BN in Figure 1.

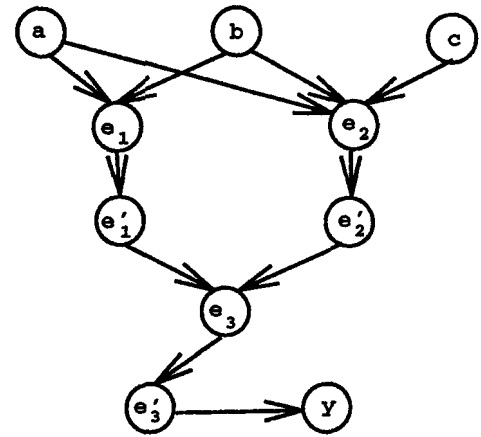

Figure 2: The BN in Figure 1 after the deputation of convergent variables.

The heterogeneous factorization denoted by the deputation BN consists of heterogeneous factors: $f_{11}(e_1, a)$, $f_{12}(e_1, b)$, $f_{21}(e_2, a)$, $f_{22}(e_2, b)$, $f_{23}(e_2, c)$, $f_{31}(e_3, e'_1)$, and $f_{32}(e_3, e'_2)$; and homogeneous factors: $P(a)$, $P(b)$, $P(c)$, $P(y|e'_3)$, $I(e_1, e'_1)$, $I(e_2, e'_2)$, and $I(e_3, e'_3)$. Note that deputy variables are regular variables by definition and deputing functions are a homogeneous factor by definition.

Also note that in the deputation BN, the combination of all the heterogeneous factors is the same as the multiplication of the conditional probabilities of all the convergent variables. The same is not true without deputation. One implication is that in a deputation BN, the joint probability equals to the combination of all the heterogeneous factors times the multiplication of all the homogeneous factors.

### 4.5 A constraint on elimination ordering

The first two steps in summing out a variable $z$ from a deputation BN can be: (1) remove from the list of heterogeneous factors all the factors that involve $z$, combine them by the general combination operator resulting in, say, $f$; and (2) remove from the list of homogeneous factor all the factors that involve $z$, combine them by multiplication resulting in, say, $g$. The next step would be to combine $f$ and $g$ by multiplication. To guarantee the correctness of doing so, deputy variables must be summed out *after* their corresponding convergent variable (Zhang and Poole 1994b).

An ordering by which variables outside $X$ is summed out is usually referred to as an *elimination ordering*. A *legitimate* elimination ordering is one where convergent variables always appear before their deputies.

The legitimacy constraint on elimination ordering can be enforced in two steps[1]. First, replace the convergent variables in $X$ with their deputies, resulting in a new list of variables $X'$. It is evident that $P(X', Y=Y_0)$ is the same as $P(X, Y=Y_0)$.

---
[1] This improvement over Zhang and Poole (1994b) is suggested by Wei Xiong.

Second, find a legitimate elimination ordering of variables outside $X'$. Such an ordering can be found by using, with minor adaptations, the maximum cardinality search heuristic (Tarjan and Yannakakis 1984) or the minimum deficiency heuristic (Bertelè and Brioschi 1972).

Note that the first step is necessary, because otherwise we will not be able to sum out the deputies of the convergent variables in $X$ due to the legitimacy constraint.

Also note that an legitimate elimination ordering contains all the variables in $Y$. Remember that irrelevant parts of factor arrays have been pruned and a regular variable $y \in Y$ is treated as a dummy variable with only one possible value $y_0$. However, a convergent variable in $Y$ still have more than one possible values.

### 4.6 An algorithm

Given a legitimate elimination ordering $\rho$ and the heterogeneous factorization of the deputation BN under discussion, $P(X', Y=Y_0)$ can be computed by using the *ICI (Inference with Causal independence) algorithm* given in the following.

Procedure ICI

1. **While** $\rho$ is not empty,
   - Remove the first variable $z$ from $\rho$.
   - Remove from the list of heterogeneous factors all the factors $f_1, \ldots, f_k$ that involve $z$, and set
   $$f = \otimes_{i=1}^{k} f_i.$$
   Let $B$ be the set of all the variables that appear in $f$.
   - Remove from the list of homogeneous factors all the factors $g_1, \ldots, g_m$ that involve $z$, and set
   $$g = \prod_{j=1}^{m} g_j.$$
   Let $C$ be the set of all the variables that appear in $g$.
   - If $k=0$, define a function $h$ by
   $$h(C-\{z\}) = \begin{cases} g(C)|_{y=y_0} & \text{if } z=y \in Y \\ \sum_z g(C) & \text{otherwise} \end{cases}$$
   Put $h$ onto the homogeneous factor list,
   - Else if $m=0$, define a function $h$ by
   $$h(B-\{z\}) = \begin{cases} f(B)|_{y=y_0} & \text{if } z=y \in Y \\ \sum_z f(B) & \text{otherwise} \end{cases}$$
   Put $h$ onto the heterogeneous factor list,
   - Else define a function $h$ by
   $$h(B \cup C - \{z\}) = \begin{cases} f(B)g(C)|_{y=y_0} & \text{if } z=y \in Y \\ \sum_z f(B)g(C) & \text{otherwise} \end{cases}$$
   Put $h$ onto the heterogeneous factor list. **Endwhile**
2. Combine all the heterogeneous factors by $\otimes$ resulting in, say, $f$.
3. Combine all the homogeneous factors by multiplication resulting in, say, $g$.
4. Multiply $f$ and $g$ and return the resulting factor.

Note that in the ICI algorithm, summing out a variable requires combining only the factors that involve the variable. This is why ICI lead to efficiency gains when causal independencies are present. More specifically, if causal independencies were ignored, summing out one variable would require combining all the conditional probabilities that involve the variable. With ICI, we combine all the *factors* that involve the variable. There are efficiency gains because the factors might contain less variables than the conditional probabilities.

In Figure 1, for instance, summing out variable $a$ would require combining $P(e_1|a, b)$ and $P(e_2|a, b, c)$ when causal independencies were ignored. Five variables participate in the process. By using ICI, on the other hand, we need to combine only $f_{11}(e_1, a)$ and $f_{21}(e_2, a)$. There are only three variables involved in the process.

## 5 Experiments

Experiments have been performed on the CPSC network (the version released in August 1994) to answer the following two questions: How much efficiency gains one can expect by making use of causal independencies? How effective the ICI algorithm is in answering queries posed to the CPSC network?

### 5.1 Efficiency gains due to causal independence

To answer the first question, we consider the task of computing the marginal probability for each variable in the CPSC network, and compare the computational costs incurred by the ICI algorithm and those incurred by clique tree propagation.

The *size* of a factor is defined to be the multiplication of the numbers of possible values of all the variables in the factor. A factor containing three binary variables, for instance, has a size of 8.

When computing the marginal probability of a variable, new factors are created. The maximum factor size is said to the *cost of the computing the marginal probability of the variable*, or simply the *cost of the variable*. If the inference algorithm used is ICI, we call it the *ICI cost of the variable*. On the other hand, if the inference algorithm used is clique tree propagation we call it the *CTP cost of the variable*.

Table 1 shows the distribution of variables according to their ICI costs. The "cost"-columns show ICI costs, while the "CNV" columns show the numbers of variables with ICI costs no larger than the ICI costs in the same rows. CNV is a shorthand for Cumulative Number of Variables. Table 2 shows the distribution of variables according to their CTP costs. Those statistics were computed from elimination orderings generated by the maximum cardinality search heuristic.

We see that the CTP costs of the variables are much larger than their ICI costs. For example, there are 194 variables with ICI costs in the range $(8, 512]$, while there are only 87 variables whose CTP costs are in the range $(8, 576]$. There are 293 variables with ICI costs in the range $(8, 786432]$, while there are only 144 variables whose CTP costs are in the same range. The number of variables with ICI costs no larger than 8 is roughly the same as the number of variables with CTP costs in the same range. Those variables are trivial in the sense that the portions of the CPSC network relevant to them are trees.

Table 1: Variable distribution by ICI costs

| cost | CNV | cost | CNV | cost | CNV |
|---|---|---|---|---|---|
| 8 | 124 | 1920 | 384 | 98304 | 412 |
| 96 | 255 | 2048 | 385 | 196608 | 414 |
| 192 | 285 | 3072 | 388 | 786432 | 417 |
| 384 | 301 | 6144 | 392 | 1179648 | 418 |
| 512 | 318 | 12288 | 397 | 3145728 | 420 |
| 768 | 330 | 36864 | 400 | 12582912 | 422 |

Table 2: Variable distribution by CTP costs

| cost | CNV | cost | CNV | cost | CNV |
|---|---|---|---|---|---|
| 8 | 123 | 1024 | 211 | $6.7 \times 10^7$ | 407 |
| 24 | 156 | 49152 | 227 | $10^8$ | 410 |
| 64 | 187 | 786432 | 270 | $2.6 \times 10^8$ | 412 |
| 128 | 200 | 1179648 | 370 | $10^9$ | 418 |
| 256 | 202 | 9437184 | 387 | $1.6 \times 10^9$ | 422 |
| 576 | 210 | 12582912 | 390 | | |

There are also 31 variables whose CTP costs are equal to or larger than the maximum ICI cost 12582912. Experiments have shown that a factor size of 12582812 is too large to handle (for SPARCclassic with 16MG memory). Thus with CTP, one would not be able to compute the marginal probabilities for those 31 variables. On the other hand, with ICI we have been able to compute the marginal probabilities for all the variables but 2.

### 5.2 Effectiveness of the ICI algorithm

To determine the effectiveness of the ICI algorithm, we first attempted to compute the marginal probability for each variable in the CPSC network. We were able to compute the marginal probabilities for all the variables except for 2. Table 3 shows the distribution of variables according the time it took to compute their marginal probabilities. The "time"-columns display the CPU time consumption in seconds, and the "CNV" columns show the number of variables whose marginal probabilities were computed in a time less than or equal to the corresponding CPU time. Those statistics were collected on a SPARCclassic workstation, which has a clock rate of 50mhz.

We see the for 396 out of the 422 variables, marginal probabilities can be computed in less than 1 second CPU time. The marginal probability of the variable `abdominal-pain-excerbated-by-meals`, whose ICI cost being 3145728, took 23 second to compute.

The variables `vomiting` and `vomiting-vomitus-normal-gastric-contents` have ICI cost 12582912. The computer ran out of memory while computing the marginal probabilities of those two variables.

Table 3: Variable distribution by CPU time

| time | CNV | time | CNV | time | CNV |
|---|---|---|---|---|---|
| 0.0120 | 118 | 0.9792 | 396 | 6.0434 | 416 |
| 0.1699 | 285 | 1.3926 | 398 | 8.3653 | 417 |
| 0.2533 | 330 | 2.8592 | 406 | 19.5035 | 418 |
| 0.5369 | 378 | 3.4681 | 409 | 20.8165 | 419 |
| 0.6996 | 388 | 4.8344 | 411 | 22.9564 | 420 |

To predict the performance of the ICI algorithm on real life queries, we computed the ICI costs of 100 randomly generated five-observation queries, of 100 randomly generated ten-observation queries, and of 100 randomly generated twenty-observation queries. The distributions of the queries according to their ICI costs are displayed in Tables 4, 5, and 6. The ICI cost of a query is defined in the same way as the ICI cost of a variable. Those statistics were computed from elimination orderings generated by the minimum deficiency heuristic, which is found to be slightly better than the maximum cardinality heuristic in our case.

Table 4: Distribution of five-observation queries by their ICI costs

| cost | CNV | cost | CNV | cost | CNV |
|---|---|---|---|---|---|
| 96 | 3 | 98304 | 75 | 1572864 | 92 |
| 1024 | 19 | 131072 | 76 | 3145728 | 94 |
| 12288 | 58 | 196608 | 81 | 8388608 | 98 |
| 18432 | 62 | 393216 | 83 | 16777216 | 99 |
| 32768 | 67 | 786432 | 89 | 25165824 | 100 |

Table 5: Distribution of ten-observation queries aby their ICI costs

| cost | CNV | cost | CNV | cost | CNV |
|---|---|---|---|---|---|
| 384 | 1 | 786432 | 71 | 8388608 | 93 |
| 6144 | 27 | 1048576 | 72 | 12582912 | 96 |
| 65536 | 44 | 2097152 | 80 | 67108864 | 98 |
| 131072 | 51 | 3145728 | 87 | 100663296 | 99 |
| 393216 | 60 | 4718592 | 89 | 402653184 | 100 |

Table 6: Distribution of twenty-observation queries by their ICI costs

| cost | CNV | cost | CNV | cost | CNV |
|---|---|---|---|---|---|
| 1536 | 2 | 294912 | 33 | 3145728 | 69 |
| 6144 | 6 | 524288 | 49 | 12582912 | 80 |
| 12288 | 8 | 786432 | 57 | $4 \times 10^8$ | 97 |
| 98304 | 24 | 1048576 | 58 | $10^9$ | 98 |
| 196608 | 30 | 1572864 | 64 | $1.6 \times 10^9$ | 100 |

Since we were able to compute the marginal probability of the variable `abdominal-pain-excerbated-by-meals` in 23 CPU seconds and the ICI cost of the variable is 3145728, we predict that the ICI algorithm is able to answer 94% of the five-observation queries, 87% of the ten-observation queries, and 69% of the twenty-observation queries.

Finally, our purpose in the experiments has been to demonstrate the benefits of making use of causal independence. As a consequence, other ideas such as zero compression has not been incorporated in the implementation. Program tracing revealed that in the arrays representing large factors, the majority of the array cells are zero. Thus the performance statistics can be much improved with zero compression.

## 6 Related work

The concept of causal independence given in this paper is a special case of the more general definition given by Heckerman (1993) and Heckerman and Breese (1994). It is also a special case of the generalized noisy-OR model proposed by Srinivas (1993).

Kim and Pearl (1983) proposed an approach for making use of causal independence in BNs which are polytrees based on a message-passing algorithm by Pearl (1988). D'Ambrosio (1994) proposed another approach for two level BNs with binary variables based on his earlier work on symbolic probabilistic inference. This paper has been concerned with general BNs.

Heckerman (1993) uses causal independence to alter the topologies of BNs in order to gain inference speedups. With the ICI algorithm, summing out one variable requires combining only those factors that contain the variable. The same is not true for Heckerman's approach.

## 7 Conclusion

This paper has described the ICI algorithm for BN inference. The algorithm exploits causal independencies to gain computational efficiency. Experiments on the CPSC network show that it is able to answer 420 of the 422 possible zero-observation queries, 94 of 100 randomly generated five-observation queries, 87 of 100 randomly generated ten-observation queries, and 69 of 100 randomly generated twenty-observation queries.

**Acknowledgement**

I thank Maclcolm Pradhan and Gregory Provan for sharing the CPSC network with me. The paper has benefited from a course project by Jin Gu and Wei Xiong. The term convergent variable was suggested by Glenn Shafer. Research was supported by HKUST grant DAG94/95.EG14.

## References

[1] U. Bertelè and F. Brioschi (1972), *Nonserial dynamic programming*, Mathematics in Science and Engineering, Vol. 91, Academic Press.

[2] B. D'Ambrosio (1994), Symbolic probabilistic inference in large BN2O networks, in *Proceedings of the Tenth Conference on Uncertainty in Artificial Intelligence*, pp. 128-135.

[3] D. Geiger, T. Verma, and J. Pearl (1990), d-separation: From theorems to algorithms, in *Uncertainty in Artificial Intelligence* 5, pp. 139-148.

[4] D. Heckerman (1993), Causal independence for knowledge acquisition and inference, in *Proceedings of the Ninth Conference on Uncertainty in Artificial Intelligence*, pp. 122-127.

[5] D. Heckerman and J. Breese (1994), A new look at causal independence, in *Proceedings of the Tenth Conference on Uncertainty in Artificial Intelligence*, pp. 286-292.

[6] F. V. Jensen, K. G. Olesen, and K. Anderson (1990), An algebra of Bayesian belief universes for knowledge-based systems, *Networks*, 20, pp. 637 - 659.

[7] J. Kim and J. Pearl (1983), A computational model for causal and diagnostic reasoning in inference engines, in *Proceedings of the Eighth International Joint Conference on Artificial Intelligence*, Karlsruhe, Germany, pp. 190-193.

[8] S. L. Lauritzen and D. J. Spiegelhalter (1988), Local computations with probabilities on graphical structures and their applications to expert systems, *Journal of Royal Statistical Society B*, 50: 2, pp. 157 - 224.

[9] S. L. Lauritzen, A. P. Dawid, B. N. Larsen, and H. G. Leimer (1990), Independence Properties of Directed Markov Fields, *Networks*, 20, pp. 491-506.

[10] J. Pearl (1988), *Probabilistic Reasoning in Intelligence Systems: Networks of Plausible Inference*, Morgan Kaufmann Publishers, Los Altos, CA.

[11] M. Pradhan, G. Provan, B. Middleton, and M. Henrion (1994), Knowledge engineering for large belief networks, in *Proceedings of the Tenth Conference on Uncertainty in Artificial Intelligence*, pp. 484-490.

[12] S. Srinivas (1993), A generalization of the Noisy-Or model, in *Proceedings of the Ninth Conference*

on *Uncertainty in Artificial Intelligence*, pp. 208-215.

[13] R. E. Tarjan and M. Yannakakis (1984), Simple linear time algorithm to test chordality of graphs, test acyclicity of hypergraphs, and selectively reduce acyclic hypergraphs, *SIAM J. Comput.*, 13, pp. 566-579.

[14] N. L. Zhang and D. Poole (1994a), A simple approach to Bayesian network computations, in *Proceedings of the Tenth Canadian Conference on Artificial Intelligence*, pp. 171-178.

[15] N. L. Zhang and D. Poole (1994b), Intercausal independence and heterogeneous factorization, in *Proceedings of the Tenth Conference on Uncertainty in Artificial Intelligence*, pp. 606-614.